# Robotic grasp detection based on Transformer

Mingshuai Dong[1] and Xiuli Yu[1]

**Abstract.** Grasp detection in a cluttered environment is still a great challenge for robots. Currently, the Transformer mechanism has been successfully applied to visual tasks, and its excellent ability of global context information extraction provides a feasible way to improve the performance of robotic grasp detection in cluttered scenes. However, the insufficient inductive bias ability of the original Transformer model requires large-scale dataset training, which is difficult to obtain for grasp detection. In this paper, we propose a grasp detection model based on encoder-decoder structure. The encoder uses a Transformer network to extract global context information. The decoder uses a fully convolutional neural network to improve the inductive bias capability of the model and combine features extracted by the encoder to predict the final grasp configuration. Experiments on VMRD datasets demonstrate that our model performs much better in overlapping object scenes. Meanwhile, on the Cornell Grasp Datasets, our approach achieves an accuracy of 98.1%, which is comparable with state-of-the-art algorithms.

**Keywords:** Grasp detection, Cluttered environment, Transformer

## 1    Introduction

The purpose of the deep learning-based grasp detection model is to identify a set of suitable grasp configurations from an input image. Currently, predicting grasp configurations of objects in cluttered or stacked scenarios remains a challenging task. Firstly, for grasp detection, it is not only to accurately predict the position of the object but also to model the pose and contour information of the object to predict the angle and opening distance of the gripper when the robot is grasping. Secondly, it is challenging to extract and map the complex and changing robot working environment features, which requires a large-scale dataset for model training. However, grasp detection datasets are very expensive to make. Therefore, we need to find an algorithm that can perfectly map the features of the input image to the grasp configuration through the training on the limited scale datasets.

In recent works [1]-[5], researchers have mainly focused on applying deep learning networks based on convolutional neural networks (CNNs) to grasping detection. They have achieved satisfactory results in the single object grasping detection task. In objects overlapping and cluttered scenes, however, grasp detection performance still has a lot of room to improve because of the complexity of its features. [6] proposed a robotic grasp detection approach based on the region of interest (ROI) in the cluttered multi-object scene. The method divides the regions of interest according to the instance information of the object and then predicts the grasp configuration of each region. [7] proposed a two-stage multi-task model, which can simultaneously realize grasp detection, object detection, and object manipulation relationship reasoning. In order to improve the reasoning efficiency of the multi-task model, [8] proposed a one-stage full convolutional neural network to predict the position, grasp configuration, and manipulation relations of different objects and achieved acceptable results. Most of the above methods use deeper networks and down-sampling mechanisms to increase the receptive field and improve the global feature extraction ability of the model so as to better predict the object grasp configuration and other attributes. However, with the increase of model depth, some valuable features will disappear, which limits the improvement of model performance.

Recently, Transformer [9] has been widely used in the field of natural language processing and computer vision due to its excellent ability to extract global context features. Transformer, with parallel sequences as input, can better convey the fusion of information across global sequences at the beginning of the model, reducing feature loss and improving feature representation ability. In computer vision, Transformer models represented by DETR [10], Deformable DETR [11] VIT [12], MVIT [13] and Swin [14] have achieved excellent performance in multiple visual tasks. In particular, the Swin-Transformer [14] model outperforms CNNs in image classification tasks and achieves start-of-the-art results. This further demonstrates the Transformer architecture's excellent feature extraction and feature mapping capabilities for specific tasks. However, Transformer has insufficient inductive bias capability compared to CNNs and requires large datasets for training [12].

In this paper, we propose a grasp detection model based on a Transformer and fully convolutional neural network. The model uses attentional mechanisms and sequence input to obtain adequate global feature representation and uses a fully convolutional neural network to enhance the inductive bias of the model so that it can be trained on limited-scale datasets with promising results. The transformer is used as the encoder of the model to extract the features of the input image, and a fully convolutional neural network is used as the decoder to construct the final grasping configuration. In addition, to evaluate our algorithm, we validated the performance of our model in the VMRD dataset [6] [15] and the Cornell grasp dataset [16]. Experimental results demonstrate that the Transformer mechanism can improve the robot's grasp detection performance in cluttering or stacking scenarios.

In summary, the main contributions of this paper are concluded as follows:

1) We propose a grasping detection model combining a Transformer and a fully convolutional neural network, which can be trained on a limited scale of grasp detection datasets and acquire satisfactory results.

2) We proved that our model achieved state-of-the-art results in cluttered or stacked scenarios on the VMRD dataset and achieved comparable results with state-of-the-art algorithms on the single-objective Cornell grasp datasets.

## 2    Related work

Robotic grasping has always been a desirable research topic in the field of robotics. Significantly, the application of deep learning technology in the field of robotic grasping makes the process of grasp detection free from manual features. And the deep learning model can predict all possible grasping configurations directly from RGB or RGB-D images. [1] was one of the earliest works that applied deep learning to grasp detection. They used the local constraint prediction mechanism to directly regression the grasp position in each image grid, thus realizing single-stage Multi-Grasp detection. However, this direct regression method is difficult to train and has inadequate robustness to the environment. Inspired by the Faster-RCNN [17] object detection algorithm, [3], [6] proposed a two-stage grasp detection method. This method transforms the grasp detection problem into the object detection problem and improves the efficiency and accuracy of grasping detection. However, the efficiency of a two-stage network is lower than that of a one-stage network. Therefore, [3] proposed a one-stage fully convolution model to improve the real-time performance of model prediction. In addition, in order to realize the robot grasping specified objects in a multi-objective environment, [6], [7], [8], [18], [19], and [20], et al. proposed to add object detection or instance segmentation branches into the grasp detection model to guide the model to recognize the categories of objects in the scene while detecting the grasp configuration. These multi-task models enhance the intelligence level of the robot's perception of the working environment.

From the optimization and improvement process of the grasp detection model, we can find that the development of grasp detection technology heavily follows the progress of computer vision technology. However, compared with object detection, the task of grasp detection is more complex. There are not only countless feasible grasping configurations but also strong angle restrictions on grasping positions. Therefore, it is necessary to find a better feature extraction method and a relational mapping model so that the robot can better model the global and local features of the object so as to generate a more reasonable grasp configuration.

Recently, Transformer [9], with its self-attention mechanism at its core, has achieved satisfactory results in natural language processing tasks. Moreover, because of its satisfactory global feature extraction ability and long sequence modeling ability, it gradually replaces CNNs and RNNs [21] in NLP and computer vision.

In the application of computer vision, researchers use convolution or patch embedding to encode visual information into sequence data to meet the input requirements of the Transformer. For example, DETR [10] and Deformable DETR [11] proposed the use of convolution operations to encode input images as sequential information; VIT [12], MVIT [13], and Swin-Transformer [14] proposed to split the image into patches as the input of the Transformer model. These Transformer based vision prediction models outperform traditional CNNs models in image classification, object detection, and image segmentation. However, the Transformer mechanism lacks the inductive bias capability inherent in CNNs [12], so the model needs pre-training on large-scale datasets to generalize well and achieve start-of-the-art performance. For grasping detection tasks, dataset making is a very expensive job; so far, there is no grasp dataset as large as ImageNet. Therefore, when the original Transformer is directly applied to the grasp detection model, the model cannot fully fit the relationship between the input image features and the grasp configurations, especially in cluttered or stacking scenes.

Different from previous work, in this paper, we propose a network structure with Swin-Transformer as the encoder and a fully convolutional neural network as the decoder for grasp detection. With this structure, the model has the ability of global feature modeling and the special ability of inductive bias of convolutional neural network, which enables the model to converge rapidly on smaller datasets. Compared with previous works, our model performs better in cluttered multi-object scenes while maintaining comparable results with start-of-the-art in single-object scenes on the basis of guaranteeing real-time performance.

## 3 Method

### 3.1 Grasp representation

Given an RGB image, the grasp detection model should detect not only the grasp position but also the grasp posture of an object. Therefore, [16] has proposed a five-dimensional grasp representation, which can simultaneously represent the position of the center point, rotation angle, and opening size of the parallel plate gripper and has been widely used in other grasp works [1]-[7]. In our work, we also use this representation. At the same time, in order to increase the representation ability in the multi-object environment and enable the robot to grasp the specified category object, we add a dimension representing the object category on this basis. Therefore, the grasp representation of our model's final output can be expressed as:

$$g = \{x, y, w, h, \theta, c\} \tag{1}$$

where $(x, y)$ is the pixel coordinates of the center point of grasp position, $w$ is the opening size of the parallel plate gripper, $h$ is the width, and $\theta$ is the angle between the closing direction of the parallel plate gripper and the horizontal direction, and $c$ is the corresponding object category of the grasp representation.

### 3.2 Overview

In this paper, the grasp detection model based on the Transformer architecture proposed by us consists of two parts, the encoder with Shifted Windows (Swin) Transformer as the component and the decoder with the convolutional neural network as the component. The overview structure is shown in Fig. 1 (A). Input an RGB image through the patch partition layer and split it into non-overlapping image regions. Each region serves as a token for Transformer input. More detailed, an image with an input size of $I = \mathbb{R}^{W \times H \times 3}$ is split into fixed-size patches $x = \mathbb{R}^{N \times (P \times P \times 3)}$ in its spatial dimension, where $N = (W \times H) / P^2$ represents the number of patches generated by the image split, and $P \times P$ is the size of each image patch. Then position embedding is added into each image patch and fed into the encoder. In the encoder, behind several rounds of attention calculated, the input image patches are mapped to three sets of feature maps with different dimensions. The decoder uses convolution operation, further extracts and fuses the features of the feature map according to the task requirements, and finally, the grasp configuration of the input image is predicted by the grasp detector. The details of the Swin-Transformer-based encoder and convolution-based decoder are described as follows.

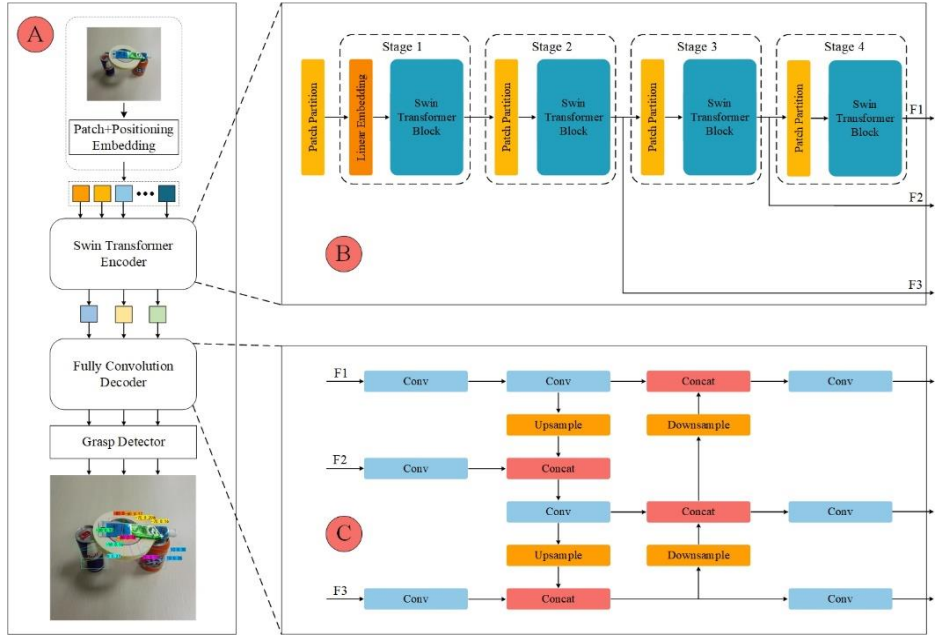

**Fig. 1.** (A). Overview architecture of our proposed model. The input RGB image is transformed into sequence data by Linear Patch embedding and added with position information, and then input into the encoder with Shifted Windows (Swin) Transformer as the component. The encoder output three different dimension features, which are input into the decoder for further feature extraction. Finally, the head of the grasp detector outputs the grasp configuration.

### 3.3 Encoder

Inspired by Swin-Transformer 14, the encoder for our model consists of four stages, as shown in Fig. 1 (B), each composed of identical Swin-Transformer blocks. The attention mechanism in each block establishes long-distance interactions across distant pixels at the beginning of the model and establishes global relevance descriptions without feature loss.

The input image token's feature $X$ is linearly transformed to derive the $Q$, $K$, and $V$ vectors. The learnable linear transformation process can be defined as follows:

$$Q = XW_Q, K = XW_K, V = XW_V \tag{2}$$

Where $W_Q$, $W_K$, and $W_V$ are learnable linear transformation matrices. On this basis, the attention calculation method between different image tokens is as follows:

$$\text{Attention}(Q, K, V) = \text{SoftMax}(\frac{QK^T}{\sqrt{d}} + B)V \tag{3}$$

Where $d$ is the dimension of $Q$ and $B$ is the relative position encoding of each token.

Because the Swin-Transformer uses a window-based attention calculation method, it has less computation than VIT. In addition, Swin-Transformer assumes the shifted windows attention mechanism to change the scope of attention and enhance the global and local feature representation capability of the model. Furthermore, the original Swin-Transformer has a hierarchical architecture that allows modeling flexibility at various scales. Therefore, to improve the ability to perceive objects of different sizes in grasp detection, we adopt a bottleneck structure and utilize three group features of different dimensions for decoding operation.

### 3.4 Decoder

The decoder uses convolution as its fundamental component to generate grasp configurations that the end-effector can operate. The purpose of using convolution as a decoder is that convolution can enhance the inductive bias of the model, thus reducing the dependence of the model training on large-scale datasets and improving the efficiency of training.

In our approach, the decoder performs feature extraction and multi-scale feature fusion for three groups of input features with different dimensions, as shown in Fig. 1 (C). The output features of each encoder are fully fused with the features of the other two dimensions after convolution, up-sampling, and down-sampling operations. The fused features are then fed into the grasp detector to predict the final grasp configuration.

In the grasp detector, we transform the grasp detection problem into pixel-level prediction, which predicts the grasp configuration, confidence, and object category of each pixel in the feature map. Finally, the optimal candidate is retained by filtering the grasp configuration through confidence score and IoU. The advantage of this approach is that only a single forward propagation can obtain the optimal grasp configuration in a global scenario.

### 3.5 Loss function

In this paper, the final prediction output of the model includes three parts: the position parameter, angle, and object category corresponding to the grasping rectangle. In addition, the angle of the grasping rectangle is predicted by the classification method. Therefore, the loss function of our algorithm consists of three parts, regression loss of grasping position, classification loss of grasping angle, and classification loss of object category.

In this paper, we employ CIoU loss function to supervise the training process of grasp position parameters. This loss function can evaluate the training process and guide the model to converge quickly by evaluating several indexes between the predicted grasp rectangle and the ground truth, such as $IoU$, the distance of the center point, and the aspect ratio. The realization process of CIoU loss function is as follows.

$$L_{grasp\_box} = L_{CIoU} = 1 - IoU + \frac{\rho^2(\mathbf{b}, \mathbf{b}^{gt})}{c^2} + \alpha \upsilon \tag{4}$$

$$\upsilon = \frac{4}{\pi^2}(\arctan\frac{w^{gt}}{h^{gt}} - \arctan\frac{w}{h}) \tag{5}$$

$$\alpha = \frac{\upsilon}{(1 - IoU) + \upsilon} \tag{6}$$

Where $\rho(\mathbf{b}, \mathbf{b}^{gt})$ is the distance between the central points of the predicted grasp rectangle and the ground truth, $c$ is the diagonal length of the smallest enclosing box covering two rectangles and $\upsilon$ represents the similarity of the aspect ratio between the predicted grasping rectangle and ground-truth, and $\alpha$ is the weight function.

We use the cross-entropy loss as the loss function of angle and object category prediction. We define the loss function of grasp angle and object category as follows:

$$L_{angle} = L_{obj\_class} = -\sum_{i=0}^{N-1} p_i^{gt} \log(p_i) \tag{7}$$

Where $N$ is the number of categories for angles or objects, $p^{gt} = [p_0^{gt}, p_1^{gt}, ..., p_{N-1}^{gt}]$ is the one-hot encoding of the sample's ground-truth; $p = [p_0, p_1, ..., p_{N-1}]$ is the prediction result of the model and represents the probability distribution of the category to which the sample belongs.

In general, the loss function of our algorithm in the training process can be defined as:

$$L_{total\_loss} = \omega L_{grasp\_box} + \beta L_{angle} + \lambda L_{obj\_class} \tag{8}$$

In the training process of the model, we set $\omega$ as 0.05, $\beta$ as 0.25, and $\lambda$ as 0.5.

# 4 Experiment set

## 4.1 Dataset

This paper utilizes the Cornell and the VMRD datasets to evaluate our proposed grasp recognition algorithm. The single-object Cornell Grasp dataset consists of 885 images of 244 different objects, and each image is labeled with multiple grasping positions of corresponding objects. We employ this dataset to evaluate the performance between our proposed algorithm and other start-of-the-art algorithms. In the VMRD dataset, each image contains multiple objects, and the dataset simultaneously labels each object with its category, grasping position, and grasping order. The VMRD dataset contains 4233 training images and 450 testing images, with 32 object categories. We use this dataset to demonstrate the performance of our model in multi-object and cluttered scenarios.

To accommodate the input of the Swin-Transformer structure and its data processing process, we preprocess the data of the input model. We first scaled the image to $(224 \times 224)$ and then rotated with the center of the image as the origin 18 times for each rotation of 20°. This data enhancement method effectively expanded the diversity of the dataset at different locations and angles and reduced the risk of overfitting in the model's training process.

## 4.2    Metric

In the single-objective scenario, we adopt a rectangular metric similar to 1-4 to evaluate the performance of our model. In the multi-objective scenario, we also consider adding the category information of the target to the assessment process. We consider that the correct grasp satisfying the rectangle metric is when a grasp prediction meets the following conditions.

1. The difference between the angle of the predicted grasping rectangle and the ground truth is smaller than 30°.

2. The Jacquard coefficient between the predicted grasping rectangle and the ground truth is more significant than 0.25.

The Jacquard index is defined as follow:

$$J(G, G^{gt}) = \frac{G \cap G^{gt}}{G \cup G^{gt}} \tag{9}$$

Where $G$ is the grasping configuration predicted by the model, and $G^{gt}$ is the corresponding ground truth.

## 4.3    Implementation details

Our algorithms are trained end-to-end on GTX 2080Ti with 11GB memory, using Pytorch as the deep learning framework. We set the batch size to 64 and the learning rate to 0.001, divided by 10 for every ten iterations. Finally, we use SGD as the optimizer for the model, with momentum set to 0.99.

## 5 Results and analysis

### 5.1 Results for single-object grasp

In this part, we use the Cornell grasp dataset to verify the performance of our proposed model in a single-objective scenario. Verification results on Cornell Grasp Dataset are demonstrated in Table 1, and Fig. 2.

**Table 1.** Accuracy of different methods on Cornell Grasp dataset.

| Author | Backbone | Input | Accuracy (%) | Speed (frame/s) |
|---|---|---|---|---|
| Lenz [16] | SAE | RGB | 75.6 | 0.07 |
| Redmon [1] | AlexNet | RG-D | 88.0 | 3.31 |
| Guo [2] | ZFNet | RGB | 93.2 | - |
| Zhou [3] | ResNet-101 | RGB | 97.7 | 9.89 |
| Fu-Jen[4] | ResNet-50 | RG-D | 96.0 | 8.33 |
| Zhang [6] | ResNet-101 | RGB | 93.6 | 25.16 |
| Liu D [5] | ResNet-50 | RGB-D | 95.2 | 21 |
| **Ours** | Swin-Transformer | RGB | **98.1** | **47.6** |

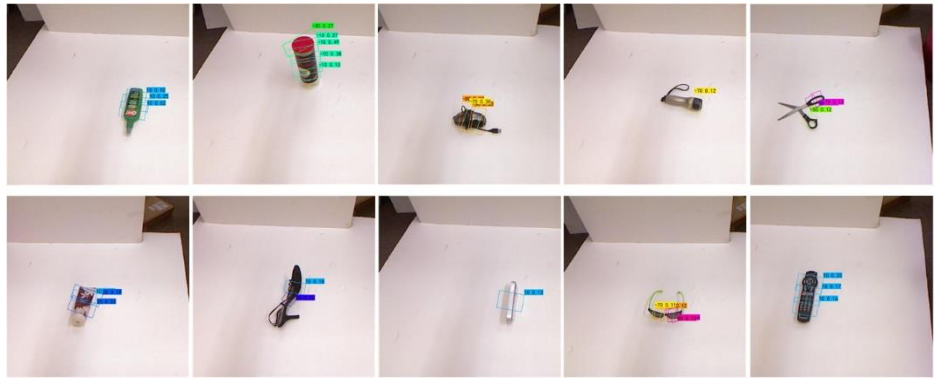

**Fig. 2.** Detection results on Cornell Grasp dataset. The color of the grasp rectangle represents the category information of the grasping angle, and the printed annotation information includes the value and confidence of the angle.

The grasp detection network based on the Swin-Transformer structure proposed by us achieves an accuracy of 98.1% with a speed of 47.6 FPS. Compared with the state-of-the-art model [3], our algorithm improves accuracy by 0.4% and improves reasoning speed five times. In more detail, we can see that compared with the two-stage model of Fu-Jen [4], Zhang [6], and Liu D [5], the one-stage model proposed by us has a better speed advantage. In addition, compared with the traditional backbone network such as AlexNet [1] and ResNet [3]-[6], our proposed network with Swin-Transformer is more comfortable in obtaining a higher detection accuracy.

### 5.2 Results for Multi-object Grasp

In order to demonstrate the grasp detection performance of our model in a complex multi-objective environment, we used the VMRD multi-objective dataset to verify our model, and the verified results are shown in Table 2 and Fig. 3. We can see that our model can accurately identify the grasp configuration of each object and its corresponding object category in the multi-object scene. Besides, our model achieves an accuracy of 80.9% on the VMRD dataset when considering categories and grasping configurations simultaneously. Compared with Zhang [6], our proposed algorithm gains 12.7% accuracy. This also proves that Transformer mechanisms can improve the model's grasp detection performance in cluttered scenarios. Compared with other models with object spatial position reasoning, such as Zhang [22] and Park [8], our model achieves higher detection accuracy, but this comparison is not rigorous. We will further improve the function of our model in future work, thus verifying the performance of the model more comprehensively.

**Table 2.** Performance summary on VMRD dataset

| Method | mAPg (%) | Speed (frame/s) |
| --- | --- | --- |
| Zhang [6], OD, GD | 68.2 | 9.1 |
| Zhang [22], OD, GD, reasoning | 70.5 | 6.5 |
| Park [8], OD, GD, reasoning | 74.6 | **33.3** |
| Ours, | **80.9** | 28.6 |

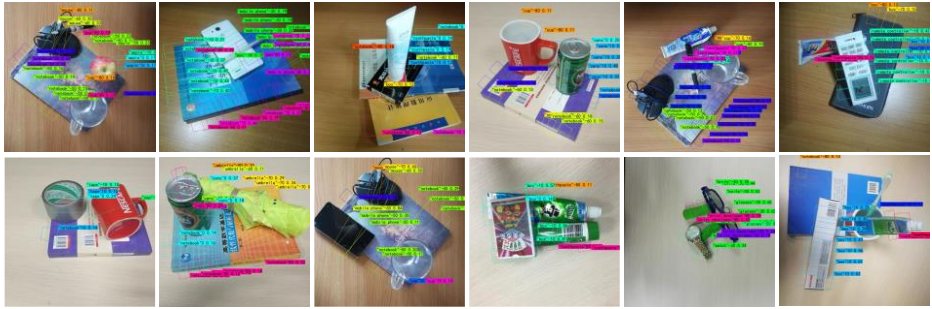

**Fig. 3.** Experimental results on VMRD dataset. The detection results of the model include the position attribute and the angle information of the grasping rectangle and the object category corresponding to the grasping configuration.

Furthermore, in order to determine the areas of our proposed model's attention, we visualized the heatmap of the graspable score, as shown in Fig. 4. From the heatmap, we can see most of the model's attention focused on the graspable parts of the object, such as the edges and center of the object. This proves that our model has accurately modeled the feature mapping from the input image features to the grasp configuration.

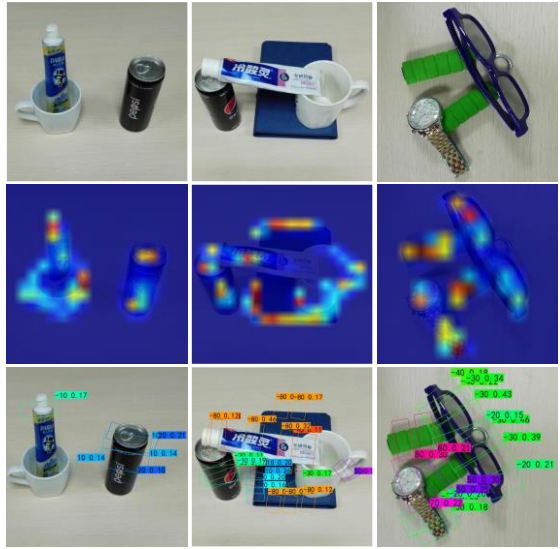

**Fig. 4.** Grasp detection results in VMRD grasp dataset. The first line is the input image of the model, the second line is the graspable score heatmap of the model, and the last line is the grasp detection result.

## 6 Conclusion and future work

In this paper, we propose a novel one-stage grasp detection algorithm based on the Transformer mechanism. Compared with other CNN-based methods and their variants, the model based on Transformer shows more flexibility for global and local feature representation and feature modeling. This attribute is particularly important for robotic grasp detection, especially in multi-object complex scenes. In addition, in order to enhance the inductive bias capability of the model and reduce the dependence of Transformer-based model training on large-scale datasets, we apply a CNN-based decoder to find reasonable feature mapping relations more quickly according to the requirements of the model. Experimental results in single-object and multi-object scenarios demonstrate that our proposed method outperforms the CNN-based models in the grasp detection performance and inference speed. In the future work, we will devote ourselves to applying the model based on the Transformer mechanism to grasp detection tasks more widely, especially in improving the accuracy of grasp detection in cluttered scenes and predicting the spatial position relationship of objects. Exploit fully the advantages of the Transformer mechanism to improve the adaptability of robots to complex features.

# References

1. Redmon J, Angelova A. Real-time grasp detection using convolutional neural networks[C]//2015 IEEE international conference on robotics and automation (ICRA). IEEE, 2015: 1316-1322.
2. Guo D, Sun F, Liu H, et al. A hybrid deep architecture for robotic grasp detection[C]//2017 IEEE International Conference on Robotics and Automation (ICRA). IEEE, 2017: 1609-1614.
3. Zhou X, Lan X, Zhang H, et al. Fully convolutional grasp detection network with oriented anchor box[C]//2018 IEEE/RSJ International Conference on Intelligent Robots and Systems (IROS). IEEE, 2018: 7223-7230.
4. Chu F J, Xu R, Vela P A. Real-world multiobject, multigrasp detection[J]. IEEE Robotics and Automation Letters, 2018, 3(4): 3355-3362.
5. Liu D, Tao X, Yuan L, et al. Robotic Objects Detection and Grasping in Clutter based on Cascaded Deep Convolutional Neural Network[J]. IEEE Transactions on Instrumentation and Measurement, 2021.
6. Zhang H, Lan X, Bai S, et al. Roi-based robotic grasp detection for object overlapping scenes[C]//2019 IEEE/RSJ International Conference on Intelligent Robots and Systems (IROS). IEEE, 2019: 4768-4775.
7. Zhang H, Lan X, Bai S, et al. A multi-task convolutional neural network for autonomous robotic grasping in object stacking scenes[C]//2019 IEEE/RSJ International Conference on Intelligent Robots and Systems (IROS). IEEE, 2019: 6435-6442.
8. Park D, Seo Y, Shin D, et al. A single multi-task deep neural network with post-processing for object detection with reasoning and robotic grasp detection[C]//2020 IEEE International Conference on Robotics and Automation (ICRA). IEEE, 2020: 7300-7306.
9. Vaswani A, Shazeer N, Parmar N, et al. Attention is all you need[J]. Advances in neural information processing systems, 2017, 30.
10. Carion N, Massa F, Synnaeve G, et al. End-to-end object detection with transformers[C]//European conference on computer vision. Springer, Cham, 2020: 213-229.
11. Zhu X, Su W, Lu L, et al. Deformable detr: Deformable transformers for end-to-end object detection[J]. arXiv preprint arXiv:2010.04159, 2020.
12. Dosovitskiy A, Beyer L, Kolesnikov A, et al. An image is worth 16x16 words: Transformers for image recognition at scale[J]. arXiv preprint arXiv:2010.11929, 2020.
13. Fan H, Xiong B, Mangalam K, et al. Multiscale vision transformers[C]//Proceedings of the IEEE/CVF International Conference on Computer Vision. 2021: 6824-6835.
14. Liu Z, Lin Y, Cao Y, et al. Swin transformer: Hierarchical vision transformer using shifted windows[C]//Proceedings of the IEEE/CVF International Conference on Computer Vision. 2021: 10012-10022.
15. Zhang H, Lan X, Zhou X, et al. Visual manipulation relationship network for autonomous robotics[C]//2018 IEEE-RAS 18th International Conference on Humanoid Robots (Humanoids). IEEE, 2018: 118-125.
16. Lenz I, Lee H, Saxena A. Deep learning for detecting robotic grasps[J]. The International Journal of Robotics Research, 2015, 34(4-5): 705-724.
17. Ren S, He K, Girshick R, et al. Faster r-cnn: Towards real-time object detection with region proposal networks[J]. Advances in neural information processing systems, 2015, 28.
18. Dong M, Wei S, Yu X, et al. Mask-gd segmentation based robotic grasp detection[J]. Computer Communications, 2021, 178: 124-130.

19. Jia Q, Cao Z, Zhao X, et al. Object Recognition, Localization and Grasp Detection Using a Unified Deep Convolutional Neural Network with Multi-task Loss[C]//2018 IEEE International Conference on Robotics and Biomimetics (ROBIO). IEEE, 2018: 1557-1562.
20. Yu Y, Cao Z, Liu Z, et al. A Two-Stream CNN With Simultaneous Detection and Segmentation for Robotic Grasping[J]. IEEE Transactions on Systems, Man, and Cybernetics: Systems, 2020.
21. Zaremba W, Sutskever I, Vinyals O. Recurrent neural network regularization[J]. arXiv preprint arXiv:1409.2329, 2014.
22. Zhang H, Lan X, Wan L, et al. Rprg: Toward real-time robotic perception, reasoning and grasping with one multi-task convolutional neural network[J]. arXiv preprint arXiv:1809.07081, 2018: 1-7.